\documentclass[pdflatex,sn-vancouver]{sn-jnl}

\usepackage{graphicx}
\usepackage{amsmath}
\usepackage{subcaption}
\usepackage{nameref}
\usepackage{xcolor}
\usepackage{booktabs}
\usepackage{hyperref}

\hypersetup{pdftitle={Astrocyte Biocomputing: Astrocytic Contributions to Neural Network Computation}, pdfauthor={Kate M. O'Neill, Ian Whitehouse, Adam McCaughan, Ryan O'Loughlin, Sonia Buckley, Wolfgang Losert}}

\newcommand{\dmgd}[0]{$\vec{\delta}$-MGD\ }

\begin{document}

\title[Astrocyte Biocomputing: Astrocytic Contributions to Neural Network Computation]{Astrocyte Biocomputing: Astrocytic Contributions to Neural Network Computation}

\author[1,2]{\fnm{Kate M.} \sur{O'Neill}}\email{oneillkm@umd.edu}
\author[1, 3]{\fnm{Ian} \sur{Whitehouse}}\email{ianjw@umd.edu}
\author[4]{\fnm{Adam N.} \sur{McCaughan}}\email{adam.mccaughan@nist.gov}
\author[4]{\fnm{Ryan} \sur{O'Loughlin}}\email{ryan.oloughlin@nist.gov}
\author*[4]{\fnm{Sonia} \sur{Buckley}}\email{sonia.buckley@nist.gov}
\author*[1,2]{\fnm{Wolfgang} \sur{Losert}}\email{wlosert@umd.edu}

\affil[1]{\orgdiv{Institute for Physical Science and Technology}, \orgname{University of Maryland}, \orgaddress{\city{College Park}, \postcode{20742}, \state{Maryland}, \country{United States of America}}}
\affil[2]{\orgdiv{Department of Physics}, \orgname{University of Maryland}, \orgaddress{\city{College Park}, \postcode{20742}, \state{Maryland}, \country{United States of America}}}
\affil[3]{\orgdiv{Department of Computer Science}, \orgname{University of Maryland}, \orgaddress{\city{College Park}, \postcode{20742}, \state{Maryland}, \country{United States of America}}}
\affil[4]{\orgdiv{Physical Measurements Laboratory}, \orgname{National Institute of Standards and Technology}, \orgaddress{\city{Boulder}, \postcode{80305}, \state{Colorado}, \country{United States of America}}}

\abstract{Conductors are critical for the performance of an orchestra even though they do not emit sound. Basic research and clinical studies in the last decade are revealing that astrocytes may be the ``conductors'' of the ``orchestra'' that is the brain network: astrocytes make up approximately 50\% of all brain cells but do not emit electrical signals. Nevertheless, they may play critical roles in important cognitive brain functions, specifically learning, memory, and information processing. Here we consider the implications of these discoveries for brain-inspired machine learning (ML), where analogy is typically drawn between neurons and their electrical interactions. We highlight specific algorithms for two key astrocyte characteristics. The first characteristic is the ability to link to and modulate thousands of synapses simultaneously. A model based on this characteristic yields faster training and makes the network more robust to noise, providing a supervised learning algorithm that can be applied to any existing neural network. The second characteristic we focus on is that the non-electrical communication among astrocytes includes slow, rhythmic dynamics. The rhythmic sharing algorithm we developed based on this premise yields rapid anomaly detection adaptation and prevents catastrophic forgetting. Overall, we find that astrocyte-derived algorithms outperform neurocentric algorithms significantly and remediate critical weaknesses of current state-of-the-art ML models.}


\maketitle

\section*{Introduction} 
\label{sec:intro}

The remarkable ability of the brain to perform efficient and fast computation is often attributed to neurons, yet these electrically-active cells comprise less than half of the brain's component cells~\cite{Von_Bartheld2016-sa}. Astrocytes, a type of glial cell that makes up approximately $50\%$ of all brain cells, have recently received increasing attention as critical and active elements in cognitive processes, such as learning and memory~\cite{Semyanov2021-rl}. In addition, astrocyte dysregulation can be associated with pathological conditions like neurodegenerative diseases~\cite{Lee2025-es, Kruyer2023-md}. 

Mechanistic studies of astrocytes have previously focused on the metabolic, mechanical, and chemical signaling of astrocytes at the cellular level~\cite{Araque1999}. A recent focus within the field is how astrocytes, via the tripartite synapse, sense and modulate synaptic activity~\cite{Noriega-Prieto2021-gg}, thereby participating in brain function. What is lacking is an understanding of how astrocytes broadly affect brain function, both at the network level \textit{in vivo} and computationally. Recent experimental work revealed astrocytes' biological roles in brain network function, including how astrocytes activate or deactivate neuronal pathways via tripartite synapse interaction, how astrocytes can stabilize neural circuit dynamics, and how astrocyte involvement may introduce irregularities to neuronal spiking ~\cite{Garcia2024-zd, Menesse2026-vo}. Here we introduce two network-level effects and show how harnessing these effects can lead to performance improvements in computational algorithms, as shown in Fig. 1.

\begin{figure}[th]
    \centering
    \includegraphics[width=\textwidth]{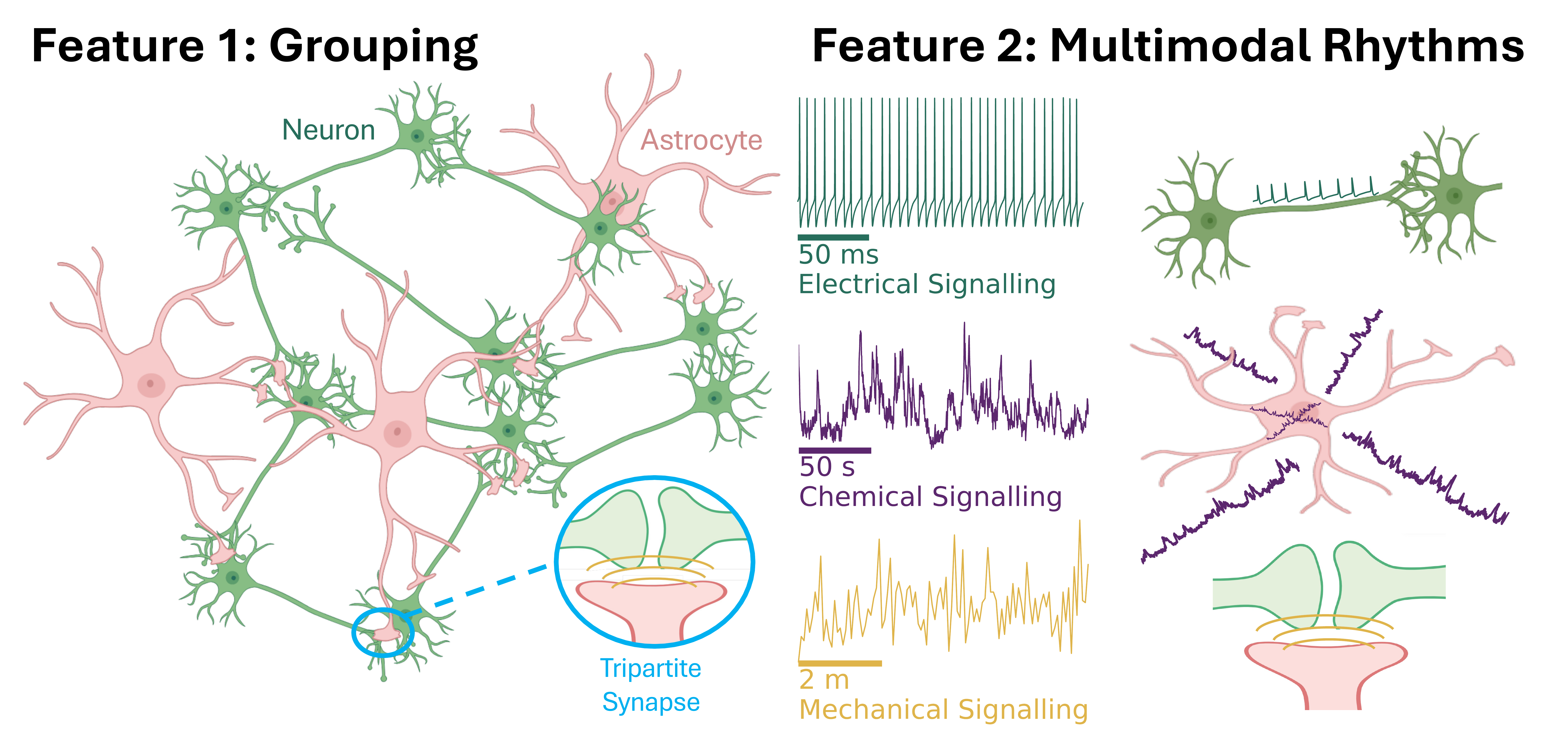}
    \caption{Figure showing the two features of astrocytes discussed in this article. \textbf{Feature 1}: Astrocytes sense and actuate groups of synapses, creating a tripartite structure at each synapse.  The intelligent grouping of different connections under one ``conductor" is central to our \dmgd algorithm.  \textbf{Feature 2}: The three different speeds of information flow between neurons and astrocytes. Foremost, the neurons communicate to each other via fast, spiking, electrical signals~\cite{Komendantov2019}. In addition, however, astrocytes generate chemical and mechanical signals at a much slower time scale~\cite{Bernardinelli2011, Gates2025-kr}.}
    \label{fig:overview}
\end{figure}

\textbf{Feature 1}: The first network-level effect is that a single astrocyte can connect to thousands of neuronal synapses~\cite{Allen2017-su} as well as to other astrocytes~\cite{Zhou2024-in, Du2022-ww}, forming coordinated groupings. As astrocytes mature, they extend complex processes that can interact with thousands of synapses (as many as 100,000 in the mouse brain~\cite{Bushong2002-dq}). Each astrocyte has a ``territory" for its processes, often referred to as a synaptic island~\cite{Halassa2007-xi}, that is non-overlapping with other astrocytes and allows each cell a distinct spatial region of influence. By interacting with thousands of neuronal synapses, astrocytes can sense and actuate broad changes in neuronal activity~\cite{Oliveira2022}. They are also able to communicate with other astrocytes within their network, or syncytium, via gap-junctional coupling \cite{Kiyoshi2019}. Computationally, this effect could have major implications in the context of training and learning. In particular, while gradient descent via backpropagation is the gold standard for training in artificial neural networks, it is suggested that the brain uses a different algorithm, given that no convincing mechanism for backpropagation in the brain has been found~\cite{lillicrap2020backpropagation}. Perturbative techniques for gradient descent, where the gradient is physically measured by perturbing individual synapses or neuron activations and measuring the change in a global cost signal, provide an alternative that is applicable to both feedforward~\cite{mccaughan2023multiplexed, Oripov2025ScalingAlgorithms} and recurrent networks~\cite{OLoughlin2025Model-FreeNetworks}. Astrocytes provide a potential mechanism for these perturbations, varying synaptic strength as a way to probe the direction of the gradient. While perturbative techniques have traditionally been thought to be too slow to be plausible alternatives to backpropagation for gradient descent~\cite{lillicrap2020backpropagation}, utilizing astrocyte-like operations can aid the training process. In the \dmgd\cite{OLoughlin2026-deltaMGD} algorithm proposed by O'Loughlin et al., using the sensing, grouping, and communication operations of astrocytes in the training process allowed the network to choose smarter perturbations, resulting in significantly more efficient learning.

\textbf{Feature 2}: The second network-level effect is that astrocytes have internal dynamics that are slow and rhythmic. It is well known that, while not electrically excitable like neurons, astrocytes' internal calcium dynamics are an important source of communication and information processing for the astrocyte network~\cite{Rusakov2025-nj, Ahmadpour2021-fl}. Release of gliotransmitters and other secreted molecules from astrocytes - which have the potential to modulate neuronal activity - is also often calcium-dependent, further emphasizing the importance of calcium signaling. As novel optical techniques have allowed for volumetric imaging over time, several groups have discovered the unique patterns of calcium oscillations within astrocytes, particularly that calcium events are not limited to increases or decreases in the soma but also appear within microdomains~\cite{Lia2021-ds, Ahmadpour2021-fl}. O'Neill et al. extend this understanding with the discovery of an additional slow rhythm within astrocytes: the mechano-chemical rhythm of the cytoskeleton~\cite{ONeill2023-ui}. In that work, astrocytes sensed and responded to changes in the external environment via dynamics of the cytoskeletal protein actin. Therefore, astrocytes appear to leverage multiple non-electrical rhythms that could affect all thousands of the synapses they interact with in a coordinated way.

The example of the rhythmic sharing algorithm highlights why these astrocytic rhythms are relevant for computation~\cite{fdc2-ljj6}. Generic reservoir computers, such as echo state networks (ESN), are efficient at predicting dynamics, but they only work well in a fixed system and cannot adapt to changes in the dynamics. These changes cause a phenomenon called ``catastrophic forgetting," where the new information causes the network to forget its original training. Rhythmic sharing mimics the oscillations that astrocytes use to signal to the neurons, allowing the network to store multiple sets of dynamics depending on the phase of the oscillations.  As new dynamics are received, the mock astrocytes encode the new information in a different subnetwork. This means that the new and original dynamics are saved separately, making the network more robust to catastrophic forgetting. Therefore, the success of this algorithm proposes an additional hypothesis for how astrocytes modulate synapses \textit{in vivo}, with astrocytes using their rhythmic action to create subnetworks where different links are synchronized with each other.

Both of the techniques we mention provide further testable hypotheses about how astrocytes might be used in the brain for computation. Studies revealing additional mechanistic insight into the mechano-chemical signaling of astrocytes would provide better input for machine learning models that incorporate astrocyte-like components. Moreover, machine learning algorithms using these principles can inform what might be happening in the brain, yielding bidirectional knowledge. Overall, we find that incorporating features of astrocyte biology into our algorithms outperforms algorithms that only include neurons and synapses. Moreover, the weaknesses we address bring the performance of ML algorithms closer to the performance of the human brain with improved anticipation and faster learning from fewer data points.

\section*{Feature 1: Astrocytes sense and actuate groups of links}
\label{sec:dmgd}

\subsection*{Biology}
Astrocytes may make up approximately $50\%$ of the brain~\cite{Von_Bartheld2016-sa} but, until the past few decades, were relegated to a supporting role in the brain, whose only computational elements were assumed to be neurons. The electrical signals of neurons—action potentials—have been extensively researched and are well described qualitatively, quantitatively, and computationally. What is significantly less well researched and less well understood is the ability of astrocytes to regulate synaptic weights and modulate neuronal communication, thereby sensing and actuating—through physical rhythms or neurotransmitters—groups of links. We propose the analogy that neurons (“the orchestra”) produce the primary computational signals, but that astrocytes (“the conductors”) direct and influence that signaling. As such, astrocytes exert much more control on the performance of brain networks than they were previously given credit for. 

The tripartite synapse introduced the idea that astrocytes may be active participants in neuronal synapses and, in turn, regulate synaptic weights~\cite{Araque1999}. Indeed, there is evidence of how astrocyte engulfment of a synapse prevents pruning, encourages synaptic maturation, and is itself a dynamic and plastic process~\cite{Allen2017-su}. The effect that astrocytes have on synapses, though, is not limited to physical interaction and influence of which synapses stay and which go. Astrocytes exert a chemical influence on neuronal communication by releasing gliotransmitters into the nearby synaptic cleft, which can influence both the presynaptic and postsynaptic neurons~\cite{Halassa2010, Volterra2005}. 

Critically, each astrocyte influences enormous numbers of synapses, potentially up to 100,000 in the rat brain (and up to 2,000,000 in the human brain)~\cite{Semyanov2021-rl}. During development, astrocytes tile themselves and carve out regions of influence that do not overlap with other astrocytes. The net result of these spatially distinct territories is astrocytes exerting broad influence over the network synchronicity and rhythmic activity of neurons~\cite{Oliveira2022}. Interestingly, if two neurons have several redundant synaptic connections that are spatially distinct, two different astrocytes could both have influence on the communication between these two neurons.
    

\subsection*{Computation}

\begin{figure}
    \centering
    \includegraphics[width=\textwidth]{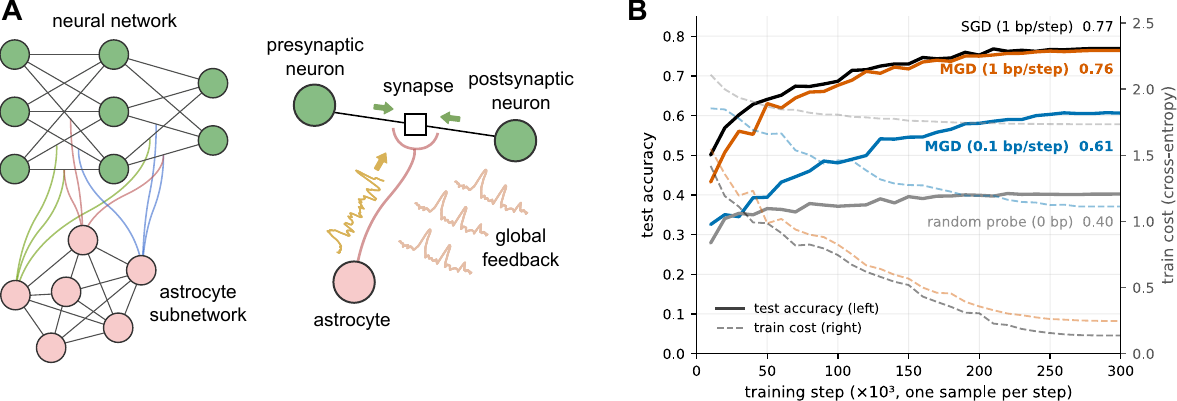}
    \caption{\dmgd~\cite{OLoughlin2026-deltaMGD} shows the influence of astrocytes on a perturbation-based, backpropagation-free learning algorithm.  \textbf{A}) Astrocytes form a distinct subnetwork that influences the computation and learning processes of the primary neuron-based neural network.  The individual astrocyte connections generate the synaptic perturbations necessary for such an algorithm. \textbf{B}) Training CIFAR-10 on a convolutional neural network with 550k parameters using the approach shown in part A, using backpropagation (bp) to estimate the astrocyte-generated perturbations. The number of backpropagations per step indicates the frequency of re-training of the astrocyte sub-network.}
    \label{fig:mgd}
\end{figure}

While astrocytes have recently been theorized to play a role in computation, one of their most important characteristics—the ability to transiently modulate synaptic strengths—remains poorly understood. Yet this same capacity connects naturally to a class of learning algorithms that depend on exactly such modulation. These perturbative algorithms compute gradients without backpropagation: instead of propagating an error signal backward through the network, they apply small perturbations to synaptic strengths and measure the resulting change in a single global, scalar error~\cite{mccaughan2023multiplexed}. Methods of this kind, including simultaneous perturbation stochastic approximation (SPSA) and multiplexed gradient descent (MGD), scale to modern machine learning problems without backpropagation~\cite{Oripov2025ScalingAlgorithms}, though they train more slowly.
The main bottleneck is that random perturbations require many samples to yield an accurate estimate of the gradient. Learning would be far more efficient with ``smart" perturbations—ones that give better gradient estimates, ideally pointing directly along the true gradient. This idea is the basis of the \dmgd algorithm~\cite{OLoughlin2026-deltaMGD}, in which an astrocyte sub-network generates perturbations for the primary network that align with the gradient far better than random ones do. The sub-network takes the class labels from the data and outputs an estimate of the bias gradient for each neuron; the perturbation applied to a given synapse is then simply the product of the presynaptic neuron's activation and the estimated bias gradient of the postsynaptic neuron. Alternatively, the astrocyte could perturb the synaptic weight directly. The resulting change in the global error is used to update the synaptic strengths. Fig. \ref{fig:mgd}A shows a schematic illustrating the algorithm on a small network. 

All of these processes are consistent with the observed behavior of astrocytes: they are touching both the pre- and post- synaptic connections, and can sense pre- and post- synaptic information, as well as information on the inputs and globally available cost/reward signals~\cite{Halassa2007-xi}. The gradient being estimated also has a low-rank structure, so many of its elements are effectively grouped together, reducing the degrees of freedom needed to construct a useful perturbation. Astrocytes are therefore a compelling candidate for this process precisely because they combine the two capabilities it requires: they can perturb synaptic strengths, and they naturally group large numbers of synapses within a single spatial territory. Fig.~\ref{fig:mgd}B shows the training performance of a variant of the MGD algorithm that directly perturbs the weights training a 550k parameter convolutional neural network to solve CIFAR-10 to the same accuracy as stochastic gradient descent. The time to solve to a given accuracy decreases with how often the perturbation direction per class is updated by an astrocyte sub-network, as illustrated by the orange and blue lines.  

\subsection*{Outlook}

\dmgd relies on the ability of astrocytes to provide estimates of the gradient of the current training input cost with respect to synaptic weights or neuron biases. Thus, if a similar mechanism is present in the brain, astrocytes must adapt their perturbations to the current network inputs. Understanding the bioplausibility of these algorithms requires increasing our understanding of the details of the mesoscale information flow that the astrocytes see. For example, questions remain about how astrocytes react to presynaptic neuron signals, the states of other neurons and synapses, and how they integrate the information from and across the synapses that they touch. Further important questions remain about the scale of that signal and how many neurons an individual astrocyte sees, whether individual astrocytes can actuate different synapses in different directions, and whether and how these actuation directions are affected by other signals in the network. The answers to these questions can form a series of rules from which new astrocyte-inspired algorithms can be tested and adapted.

The preliminary studies have shown that how effectively these astrocyte networks can speed up training over random perturbations depends on the problem, the network topology, and the rules of the algorithm. Further study is required to unlock the full potential of astrocyte sub-networks and could lead to significantly faster backpropagation-free gradient descent algorithms.

\section*{Feature 2: Astrocytes have mechano-chemical rhythms}
\label{sec:rs}

\subsection*{Biology}

Cells' shapes and movement are controlled by an actin cytoskeleton, a web of thin filaments. Previous research has shown that, in addition to maintaining shape and allowing movement, the cytoskeleton contributes to and is necessary for cell function. For example, microglia (the brain's immune cells) use filopodia to monitor brain health and homeostasis~\cite{Bernier2019-so}.

Within astrocytes, the cytoskeleton is known to play a role in the development of the tripartite synapse~\cite{Badia-Soteras2025-ns}. Following previous research showing that the actin cytoskeleton in various cells displays oscillatory/wave-like motion, O’Neill et al. discovered that astrocytes' actin cytoskeleton oscillates in response to chemical and physical stimuli~\cite{doi:10.1126/sciadv.aay7682, Sun2015-dx, Miao2019-sr, Vicker2002-rr, Roy2016-wt, ONeill2023-ui}. These dynamics were concentrated into actin ``hot spots", which are enhanced by the presence of neurons.

This body of work suggests that astrocytes use these cytoskeletal rhythms to communicate external stimuli to neurons, as it is known that mechanical pressure affects synapse firing~\cite{Ucar2021-iw}. In addition, other works have found that the disruption of actin dynamics leads to a breakdown in the cells' calcium signaling ability and also plays a key role in neurodegenerative disease, supporting the hypothesis that the slow communication between neurons and astrocytes is necessary for cognition~\cite{Gates2025-kr, Kc2026-pa}.


\subsection*{Computation}

\begin{figure}
    \centering
    \includegraphics[width=\textwidth]{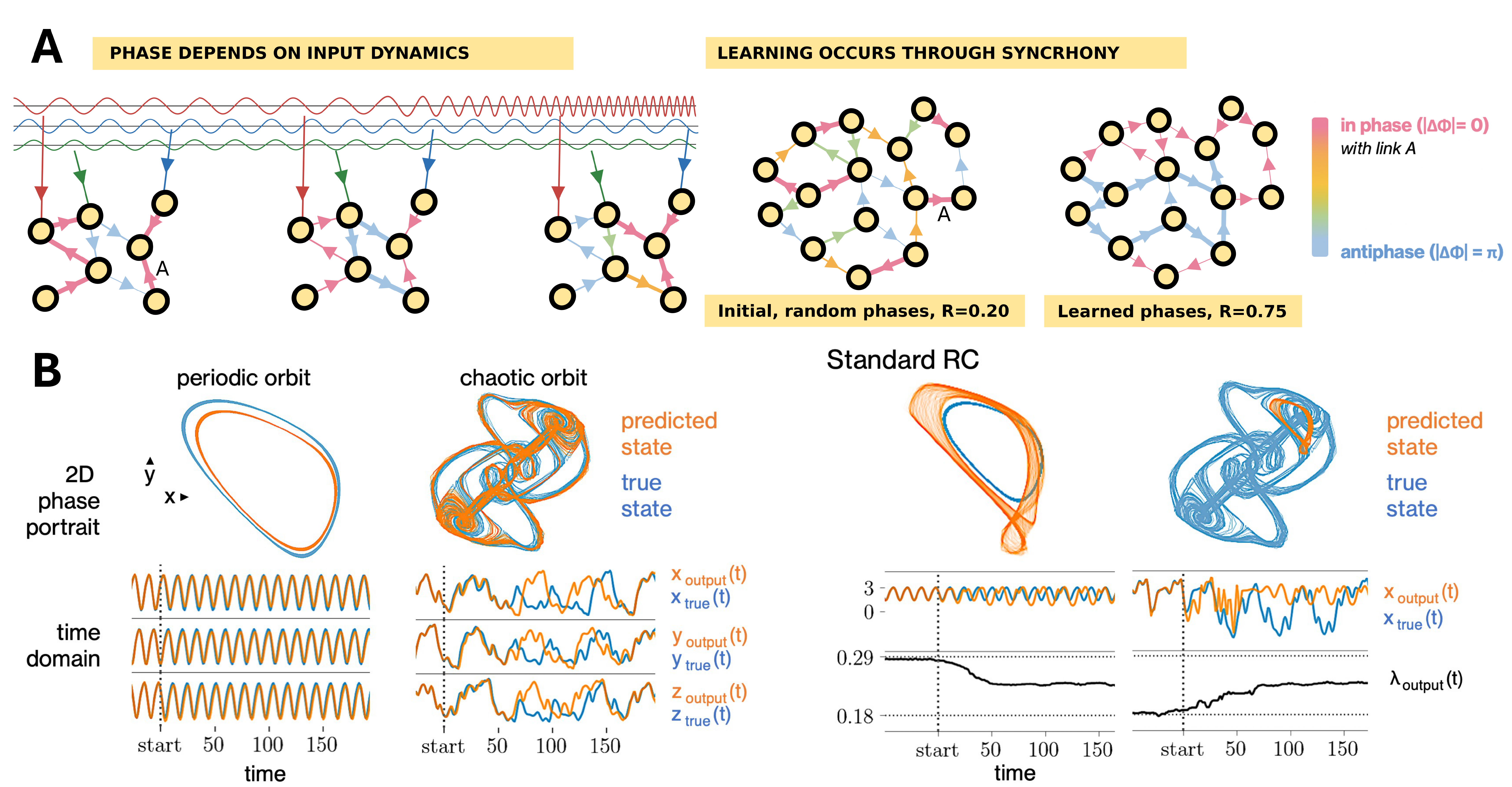}
    \caption{Rhythmic sharing algorithm by Kang and Losert, adapted from previous publications~\cite{fdc2-ljj6}.  \textbf{A}: An overview of the algorithm highlighting how the astrocytes change which nodes are synchronized as the input data changes.  \textbf{B}: An example of how this algorithm improves on a generic echo state network.  The astrocytes allow the algorithm to remember multiple sets of dynamics while the generic ESN learns a combination of the two.}
    \label{fig:rs}
\end{figure}

Astrocytic communication through actin dynamics is significantly slower than neuronal synapses. However, recent work by Yang et al., which added astrocyte-inspired units to spiking neural networks, shows that these ``slow nodes" cannot act alone~\cite{Yang2026-uo}. This work found that balancing the number of astrocytes and neurons maximized the performance of the system, furthering the hypothesis that astrocytes are the ``conductors" and that both are necessary within the brain.

Another work that explored this was Rhythmic Sharing, an algorithm by Kang and Losert, finding that the combination of slower nodes and rhythmic behavior unlocks performance improvement~\cite{fdc2-ljj6}. The algorithm adds astrocyte-inspired oscillations to a reservoir computing network. As the dynamics of the input data change, the ``astrocytes" adjust the phases of the oscillations, leading to different nodes being in-phase with each other. Kang and Losert showed this behavior halted catastrophic forgetting because it allowed the network to retain knowledge of one set of dynamics despite being shown new dynamics, which the generic reservoir lacked the ability to do.  An overview of the algorithm and this ability are highlighted in Fig.~\ref{fig:rs}.

In both works, astrocytes have a ``mesoscale'' overview of neuronal activities, using this perspective to sense the input data and to improve the neurons' performance on the computational task, which we argue is central to the algorithms' success.


\subsection*{Outlook}
\begin{table}
    \begin{tabular}{l|rrr}
    \toprule
    \textbf{Datasets:} & \textbf{NASA C-MAPSS} \cite{4711414} & \textbf{SWaT} \cite{swat} & \textbf{WADI} \cite{wadi} \\
    \midrule
    Previous SOTA F1         & 0.956 \cite{somma2025anomalydetectioncomplexdynamical} & 0.811 \cite{bdm} & 0.377 \cite{bdm} \\
    Best detector F1 w/ RS   & 0.831 & 0.871 & 0.694 \\
    Best detector F1 w/o RS  & 0.676 & 0.792 \cite{nsibf} & 0.318 \cite{nsibf} \\
    $\Delta$ from RS         & $+0.155$ & $+0.079$ & $+0.376$ \\
    \bottomrule
\end{tabular}
    \caption{Results of applying the astrocyte-inspired concept drift detection from Whitehouse et al. to the NASA C-MAPSS, SWaT, and WADI datasets~\cite{Whitehouse2026-gw, 4711414, swat, wadi}.  As shown, the algorithm improves the performance of the detectors, including SOTA results from previous works~\cite{nsibf, bdm}.}
    \label{tab:SWaT_wadi}
\end{table}

Analyzing astrocytes through their slow, rhythmic, physical action on synapses has already yielded interesting results. One avenue for this is the application of the rhythmic sharing algorithm to concept drift detection~\cite{Whitehouse2026-gw}. This work harnesses the synchrony of astrocytic rhythms in the network to detect changes in the input data, similar to how astrocytes change their oscillations based on different input stimuli~\cite{ONeill2023-ui}.  Table~\ref{tab:SWaT_wadi} highlights this success.

The finding that astrocytes communicate with neurons through slow oscillations raises additional questions that we expect will be fruitful for further study. For example, a single astrocyte interacts with upwards of $1{,}000$ synapses through the same number of endfeet, each capable of mechanical oscillations. Little is known about how  the observed oscillations at endfeet are communicated within a cell, raising the possibility of intra-astrocyte rhythms or other methods of communication. Additionally, more research is necessary to understand how astrocytes communicate with each other and whether these oscillations are communicated to or shared by other astrocytes.


\section*{Discussion} 
\label{sec:conclusions}

In this paper we propose a conductor-orchestra analogy to describe the role astrocytes play in the brain. In this analogy, neurons are the orchestra: they are fast, send electrical signals, and are the primary compute substrate. In contrast, astrocytes are the conductors: they are slow, but they control which neurons link together at different points in the piece. This analogy shows how astrocytes, which make up approximately half of the cells in the brain~\cite{Von_Bartheld2016-sa}, are essential to computation in the brain, despite being largely ignored in neuromorphic computing and machine learning.

One way that astrocytes might enable learning within the brain is by grouping neuronal cells. In the brain, astrocytes form tiled, non-overlapping territories with up to $\sim 10^5$ synapses in each of their domains~\cite{Bushong2002-dq, Halassa2007-xi}. At these connections between neurons, the astrocytes take part through the tripartite synapse, where they read and modulate pre- and post-synaptic electrical activity~\cite{Noriega-Prieto2021-gg, Allen2017-su}. We explore this through the \dmgd algorithm~\cite{OLoughlin2026-deltaMGD}, where an astrocyte subnetwork groups links between artificial neurons and provides gradient-aligned perturbations. This approach exploits the low-rank structure of the gradient vector, which maps directly to the grouping based on each astrocyte node's territories. We test the algorithm on the CIFAR-10 image classification task, where our neuromorphic algorithm approaches the accuracy of stochastic gradient descent, and we find that the time the curve takes to reach a set accuracy depends on how often the astrocyte network updates (Fig.~\ref{fig:mgd}).

Astrocytes also produce slow rhythms that might enable adaptation to different computational challenges. Astrocytes communicate through actin and chemical rhythms on seconds- or minutes-long timescales~\cite{ONeill2023-ui, Rusakov2025-nj, Lia2021-ds}, with stimulus-responsive actin hot-spots enhanced by neurons~\cite{ONeill2023-ui}. Other experiments have shown that disruptions of this actin rhythm lead to calcium activity breakdown and degenerative neural pathologies~\cite{Gates2025-kr, Kc2026-pa}. In our rhythmic sharing algorithm~\cite{fdc2-ljj6}, astrocyte-inspired dynamics partition a reservoir into context-specific subnetworks, which appears to mitigate catastrophic forgetting. The synchrony between nodes in these subnetworks also acts as an emergent drift signal, improving the performance of drift-detection algorithms~\cite{Whitehouse2026-gw} (Table~\ref{tab:SWaT_wadi}). Finally, Yang et al.~\cite{Yang2026-uo} show that performance peaks at a balanced ratio of astrocytes and neurons, as found in the brain, so these slow nodes are not capable of computation without the neurons.

Both algorithms discussed have a shared idea at their base: astrocyte-inspired elements provide a slow, mesoscale controller that senses and actuates groups of fast-acting nodes. While one algorithm groups by spatial territory and the other groups in time, we propose a unified theory that the ability of astrocytes to do both is central to computation within the brain. We propose that building these astrocyte-inspired layers could alleviate many of the current challenges in machine learning: continual learning without catastrophic forgetting, unsupervised adaptation to context change, and data inefficiency.

To continue building this theory, we propose additional questions inspired by our work for both computational models and the brain. The questions about real astrocytes inspired by our algorithms are ``do astrocytes encode input-conditioned signals?'', ``can one astrocyte actuate different synapses in different directions?'', and ``do endfoot oscillations sync within a single cell or across multiple?''. The questions inspired by the brain we plan to address in algorithmic implementations are ``how does the information flow to real astrocytes constrain the algorithms we can develop?'' and ``how does the behavior of and advantage conferred by astrocyte-based nodes depend on different problems or topologies?'' In our recent work~\cite{whitehouse2026quantumerrorcorrectionbiological}, we frame neural error handling as redundant encoding on a low-dimensional codespace, with syndrome-like mismatch signals driving fast correction plus slower adaptive updates, which directly ties to the astrocyte rhythms we discuss here. Therefore, a natural next step is to extend our conductor framing to fault tolerance or quantum error correction decoder design.

In conclusion, we argue that neurocentric models in neuromorphic computing and machine learning omit a necessary and powerful computational layer, and we show that two astrocyte-based features have already yielded concrete algorithmic gains. We argue that joint experimental and computational work is necessary to understand and benefit from this new paradigm.

\section*{Declarations}
\label{sec:decl}

\bmhead{Acknowledgments}

The authors acknowledge Sylvester Gates III (University of Maryland) and Ariel Fitzmorris (University of Maryland) for their assistance with preparation of the figures in this manuscript.

This collaboration was initiated at Aspen Center for Physics, which is supported by National Science Foundation grant PHY-2210452.

This study was funded in part by the Air Force Office of Scientific Research Biophysics Program [Grants \emph{FA9550-22-1-0405} and \emph{FA9550-25-1-0405}]. This material is partially based upon work supported by the Air Force Office of Scientific Research under award number \emph{FA9550-25-C-B010} in the amount of $\$138{,}200$. The funders played no role in study design, data collection, analysis and interpretation of data, or the writing of this manuscript. 

\bmhead{Author Contributions}

SB, ROL and ANM prepared Section~``\nameref{sec:dmgd}'' and contributed to the simulations on $\vec{\delta}$-MGD. 
IW and WL prepared Section~``\nameref{sec:rs}''.
KMO prepared the biology sections.
All authors developed the integrated perspective. 

\bmhead{Competing Interests}
Certain authors may be named on a pending patent application. Inquiries regarding licensing may be directed to the NIST Technology Partnerships Office at \href{mailto:licensing@nist.gov}{licensing@nist.gov} and UM Ventures at \href{mailto:umdtechtransfer@umd.edu}{umdtechtransfer@umd.edu}. WL is founder of \href{https://astrocomputing.ai/}{Astrocomputing.AI}. The authors declare no other competing financial or non-financial interests. 

\bibliographystyle{plain}
\bibliography{bib}

\end{document}